\documentclass[runningheads]{llncs}
\usepackage[T1]{fontenc}
\usepackage{graphicx}
\usepackage{algorithm}
\usepackage{caption}
\usepackage{subcaption}
\usepackage[algo2e]{algorithm2e}
\usepackage{xcolor}
\usepackage{pdfpages}
\usepackage{algpseudocode}
\usepackage{multirow}
\begin{document}
\title{Genetic Engineering Algorithm (GEA): An Efficient Metaheuristic Algorithm for Solving Combinatorial Optimization Problems}


%
\titlerunning{Genetic Engineering Algorithm (GEA)}
%
\author{Majid Sohrabi\inst{1,2}\orcidID{0000-0003-3695-604X} \and Amir M. Fathollahi-Fard\inst{3}\orcidID{0000-0002-5939-9795} \and Vasilii A. Gromov\inst{1}\orcidID{0000-0001-5891-6597}}
\authorrunning{Sohrabi et al.}
%

\institute{School of Data Analysis and Artificial Intelligence, Faculty of Computer Science, HSE
University, 11 Pokrovsky Boulevard, 109028, Moscow, Russian Federation
\and
Laboratory for Models and Methods of Computational Pragmatics, HSE University, Moscow, Russia Federation
\\ \email{(msohrabi@hse.ru)}\\
\email{(stroller@rambler.ru)}\\
\and
School of Management Sciences, Université du Québec à Montréal, B.P. 8888, Succ. Centre-ville, Montréal, QC, H3C 3P8, Canada
\\ \email{(fathollahifard.amirmohammad@courrier.uqam.ca)}\\}

%
\maketitle              
\begin{abstract}
Genetic Algorithms (GAs) are known for their efficiency in solving combinatorial optimization problems, thanks to their ability to explore diverse solution spaces, handle various representations, exploit parallelism, preserve good solutions, adapt to changing dynamics, handle combinatorial diversity, and provide heuristic search. However, limitations such as premature convergence, lack of problem-specific knowledge, and randomness of crossover and mutation operators make GAs generally inefficient in finding an optimal solution. To address these limitations, this paper proposes a new metaheuristic algorithm called the Genetic Engineering Algorithm (GEA) that draws inspiration from genetic engineering concepts. GEA redesigns the traditional GA while incorporating new search methods to isolate, purify, insert, and express new genes based on existing ones, leading to the emergence of desired traits and the production of specific chromosomes based on the selected genes. Comparative evaluations against state-of-the-art algorithms on benchmark instances demonstrate the superior performance of GEA, showcasing its potential as an innovative and efficient solution for combinatorial optimization problems.

\end{abstract}
\section{Introduction}
Combinatorial optimization problems belonging to the class of NP-hard ones pose significant challenges in various domains, requiring efficient algorithms to find optimal or near-optimal solutions. Genetic Algorithms (GAs) ~\cite{ref1} have emerged as a popular choice due to their ability to explore diverse solution spaces, adapt to changing dynamics, and provide heuristic search. However, the limitations of GAs, including computational complexity, premature convergence, lack of problem-specific knowledge, and the need for parameter tuning, motivate the search for innovative approaches to enhance their efficiency ~\cite{ref2}. These drawbacks encourage our attempts to redesign GA using the genetic engineering concept to be highly efficient for solving combinatorial optimization problems. 

According to the literature on metaheuristic algorithms, GAs stand as the earliest population-based algorithms prioritizing the discovery of satisfactory solutions within a reasonable computational timeframe, rather than exclusively pursuing optimality~\cite{ref3}. While the field boasts an array of novel metaheuristic algorithms~\cite{ref4}--~\cite{ref18}, it is worth noting that the literature on GAs is exceptionally rich, featuring numerous research contributions that introduce various GA variants equipped with advanced genetic programming and engineering techniques~\cite{ref19}--~\cite{ref27}. For instance, Gero and Kazakov~\cite{ref28} conducted a study focusing on the identification of useful genetic material while minimizing the presence of harmful genetic components, leading to the proposition of a novel GA. Kameya and Prayoonsri~\cite{ref29} introduced a GA-based approach grounded in pattern recognition to identify essential patterns within favorable chromosomes and protect them from undesirable crossovers. Ding et al.~\cite{ref30} delved into the integration of a back-propagation (BP) neural network with GA. Liang et al.~\cite{ref31} proposed a suite of adaptive elitist-population strategies that found application within the GA framework.

Additionally, Dasgupta et al.~\cite{ref32} integrated a load-balancing strategy for cloud computing with GAs. Elsayed et al.~\cite{ref33} enhanced GAs with a novel multi-parent crossover operator. Peng and Li~\cite{ref34} put forth an improved DV-Hop algorithm based on GAs, while Askarzadeh~\cite{ref35} explored memory-based GAs. Reddy et al.~\cite{ref36} developed a hybrid GA infused with fuzzy logic, and Fathollahi-Fard et al.~\cite{ref37} proposed a hybrid of GA with other innovative metaheuristics. Furthermore, Fathollahi-Fard et al.~\cite{ref38} devised a revised non-dominated sorting genetic algorithm by introducing novel search operators. Last but not least, Kolaee et al.~\cite{ref39} introduced a local search-based non-dominated sorting genetic algorithm tailored to solving routing problems within the tourism industry. However, none of the studies reviewed thus far have proposed the introduction of new search operators grounded in a diverse array of methods aimed at isolating, purifying, inserting, and expressing new genes within existing GA chromosomes, as we have undertaken in this study.

Recently, a wide range of population-based algorithms has been proposed to address challenging optimization problems. Examples include Cuckoo Search (CS) ~\cite{ref4}, Whale Optimization Algorithm (WOA) ~\cite{ref5}, Sine Cosine Algorithm (SCA) ~\cite{ref6}, Harris Hawks Optimization (HHO) ~\cite{ref7}, Squirrel Search Algorithm (SSA) ~\cite{ref8}, Red Deer Algorithm (RDA) ~\cite{ref9}, Sparrow Search Algorithm (SSA) ~\cite{ref10}, Capuchin Search Algorithm ~\cite{ref11}, Aquila Optimizer (AO) ~\cite{ref12}, Chameleon Swarm Algorithm (CSA) ~\cite{ref13}, Aptenodytes Forsteri Optimization (AFO) ~\cite{ref14}, Dung Beetle Optimizer (DBO) ~\cite{ref15}, Beluga Whale Optimization (BWO) ~\cite{ref16}, and others. However, it is essential to note that the ``No Free Lunch'' theorem ~\cite{ref17} suggests that no metaheuristic algorithm can outperform others for all optimization problems. Hence, there is a constant demand for the development of new metaheuristic algorithms that exhibit improved performance across different problem domains ~\cite{ref18}.

In this paper, we propose a new metaheuristic algorithm, the Genetic Engineering Algorithm (GEA), inspired by genetic engineering concepts. Genetic engineering encompasses a diverse range of methods used to isolate, purify, insert, and express new genes based on existing ones, resulting in the emergence of desired traits and the production of specific chromosomes based on the selected genes. By drawing parallels from this field, we aim to redefine the optimization process and overcome the limitations inherent in traditional GAs. The techniques used in GEA enable more precise manipulation of the optimization process, leveraging problem-specific insights and reducing randomness in mutation and crossover operations. By introducing the concept of gene manipulation within the population, GEA aims to enhance the exploration and exploitation of the solution space, leading to improved convergence and solution quality. To evaluate the effectiveness of GEA, we conduct extensive experiments on a set of benchmark instances and compare its performance against state-of-the-art metaheuristic algorithms. The results demonstrate the superior performance of GEA in terms of convergence speed, solution quality, and robustness. This signifies the potential of GEA as a novel and efficient approach for solving combinatorial optimization problems.

The rest of the paper is organized as follows: Section 2 presents the main inspiration of our GEA based on the genetic engineering concept. Section 3 studies the design and implementation details of the proposed GEA based on genetic engineering operators. Section 4 presents the experimental setup and discusses the comparative results with other algorithms. Finally, Section 5 concludes the paper, by emphasizing the significance of GEA as an innovative solution for efficient combinatorial optimization problems and outlining potential directions for future research.

\section{Inspiration}

Genetic engineering (GE) has transitioned from speculative fascination to a groundbreaking reality with wide-ranging applications. This methodology exhibits significant potential in disease treatment, exemplified by cancer immunotherapy~\cite{ref19}, and the utilization of CRISPR technology to eliminate the HIV virus from infected cell genomes, offering prospects for a cure~\cite{ref20}. It extends to the realm of human genetics, impacting characteristics in newborns~\cite{ref21}, and holds promise in agriculture for producing high-yield crops~\cite{ref22}. For instance, the development of Golden Rice, engineered to combat vitamin A deficiency and prevent blindness worldwide, underscores GE's potential~\cite{ref23}.

Dominant chromosomes wield a pivotal influence on genetics, shaping specific trait expressions. Through meticulous manipulation, scientists can surpass or enhance genetic characteristics for desired outcomes. Precision in specifying dominant chromosomes related to plant yield, for instance, has led to high-yield species addressing global challenges such as climate change, pollution, and food shortages~\cite{ref24}.

Directed mutations entail precise DNA alterations in organisms to induce advantageous changes. Researchers can introduce specific mutations in genes to promote beneficial traits or suppress harmful ones. This approach finds applications in medicine, agriculture, and environmental conservation. By directing mutations in disease-causing genes, scientists are developing innovative treatments for genetic disorders like cystic fibrosis and muscular dystrophy. Achieving this involves two critical steps: identifying crucial genes for enhancing traits and carefully manipulating non-informative genes to achieve desired outcomes~\cite{ref23}.

Desired genes possess specific qualities or capabilities that scientists aim to introduce into an organism. These genes enhance crop resistance in agriculture, contributing to sustainable farming and increased food production. In medicine, they hold the key to curing genetic disorders~\cite{ref25}. Targeting desirable genes through mutation takes genetic engineering to a new level of precision. Techniques like CRISPR-Cas9~\cite{ref26} enable precise gene editing, allowing correction of mutations responsible for diseases like sickle cell anemia or Huntington's disease, offering hope to millions~\cite{ref24}.

Gene injection, another innovative approach, delivers therapeutic genes directly into the body to treat or prevent diseases. This method holds promise for conditions like cancer and cardiovascular disorders. By injecting genes producing therapeutic proteins, scientists can enhance natural defense mechanisms, stimulate tissue regeneration, or target cancer cells. Gene injection therapy represents a powerful tool in personalized medicine~\cite{ref25}.

In conclusion, genetic engineering is a groundbreaking reality with diverse applications in medicine, agriculture, and environmental conservation. Through dominant chromosomes, directed mutations, and desired gene identification, scientists shape genetics for high-yield crops and disease treatment. Techniques like precise gene editing and gene injection therapy enable unprecedented precision and customization. As genetic engineering advances, it promises to revolutionize various domains, ushering in an era of personalized medicine and sustainability. Building on these GE techniques, this paper proposes a novel metaheuristic algorithm, GEA.


\section{Proposed GEA}

\begin{figure}
\centering
\includegraphics[width=1\columnwidth]{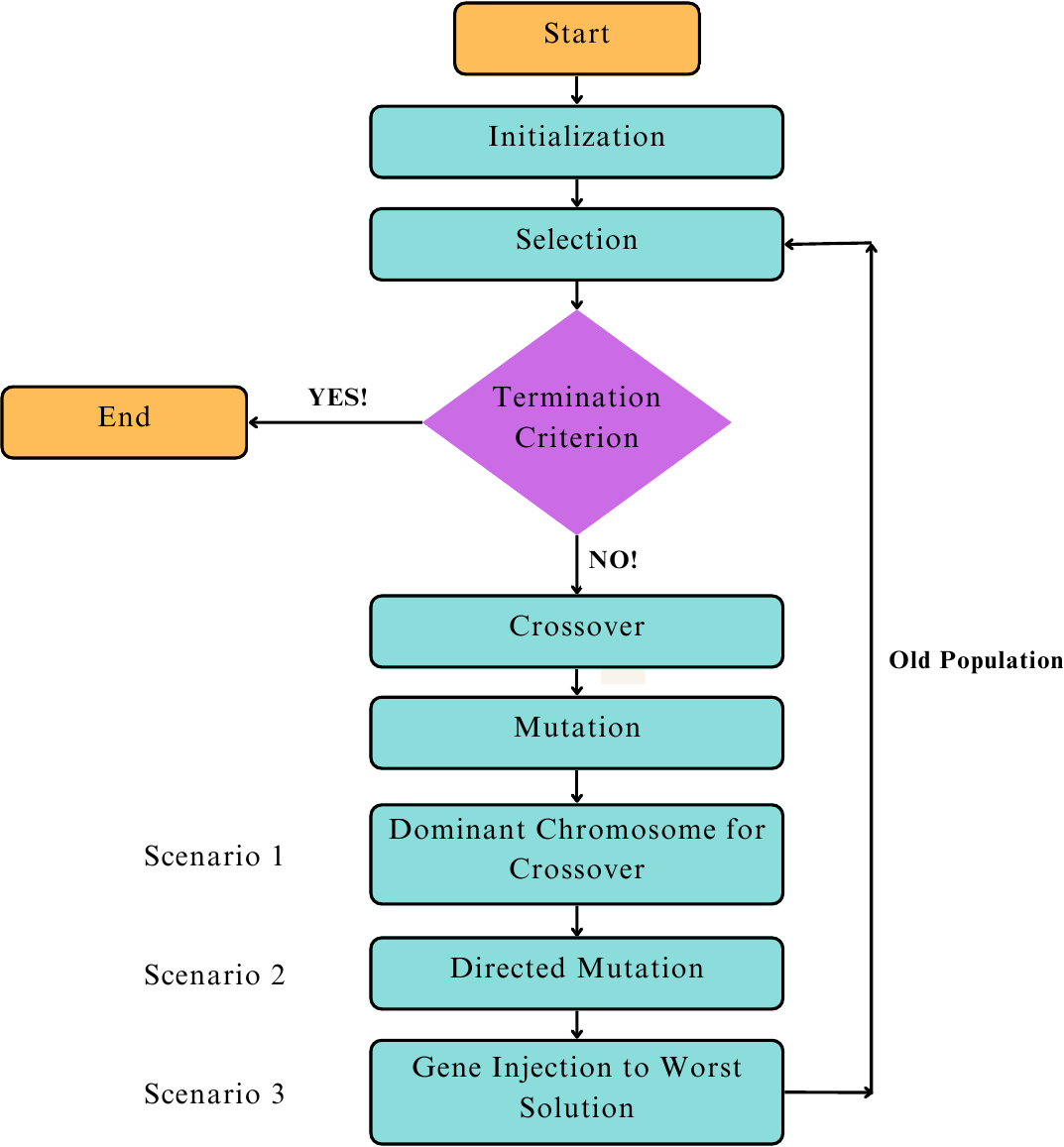}
\caption{Flowchart for the proposed GEA.}
\end{figure}

While the GA~\cite{ref1} is a well-established evolutionary algorithm that commonly employs classical mutation and crossover operators, this study proposes a novel approach by incorporating GE techniques. The overall flowchart for the proposed GEA is shown in Figure 1. One can skip any operator from Crossover to Gene Injection to customize the algorithm for different purposes and check the performance of the method with partial operators.  The GEA similar to other meta-heuristics starts with an initial population that is the counterpart of this method. This algorithm encompasses a diverse range of methods used to isolate, purify, insert, and express specific genes within a host organism, ultimately leading to the emergence of desired traits and the production of specific chromosomes based on the selected genes. The flowchart of the proposed GEA is shown in Figure 1. 

After generating the initial population, all individuals will be evaluated based on the specific fitness function. It is worth noting that each problem is defined by a specific and unique fitness function to present a solution in combinatorial optimization. We can classify different integer programming problems such as vehicle routing optimization, flow-shop scheduling problems, knapsack problems, and facility location planning as combinatorial optimization problems ~\cite{ref18}. The chromosome definition or the solution presentation in each type of combinatorial optimization problem is different. For example, the chromosome in routing optimization is defined as the sequence of visits ~\cite{ref2}. In flow-shop scheduling problems, the chromosome is considered as the sequence of a set of jobs on machines ~\cite{ref18}. In addition, the solution definition in facility location planning and knapsack problems is defined by 0-1 or binary variables ~\cite{ref3}. In our GEA, the examples for explaining the search operators are based on a binary chromosome where each gene may be zero or one. In this new metaheuristic algorithm, in addition to mutation and crossover operators, we have three genetic engineering operators as three scenarios explained as follows:

\underline{Senario 1.}
\textbf{Finding Dominant Chromosome (most repeated genes):} The first scenario of GEA focuses on identifying the dominant chromosome by considering a percentage, denoted as $p\%$, of the best individuals in the population. The value of $p$ is initially defined by the user and can be optimized based on the specific problem at hand. A chromosome is deemed dominant if it possesses the highest number of repeated genes among the best $p\%$ of individuals. The process of identifying the dominant genes and constructing the dominant chromosome is outlined by Equations (1) and (2). To provide clarity on this operation, Figure 2 illustrates an example of finding the dominant chromosome from the elite population. Additionally, the pseudocode for this operation is presented in Algorithm 1.

\begin{equation}
    RM_i = [\Sigma_{j=1}^{M} gene_j]
\end{equation}

\begin{equation}
    DC = max(RM)
\end{equation}

Where M, RM, and DC represent the number of individuals in $p\%$ of the population, repetition matrix, and dominant chromosome respectively.

\setlength{\textfloatsep}{0pt}
\begin{figure}
\centering
\includegraphics[width=1\columnwidth]{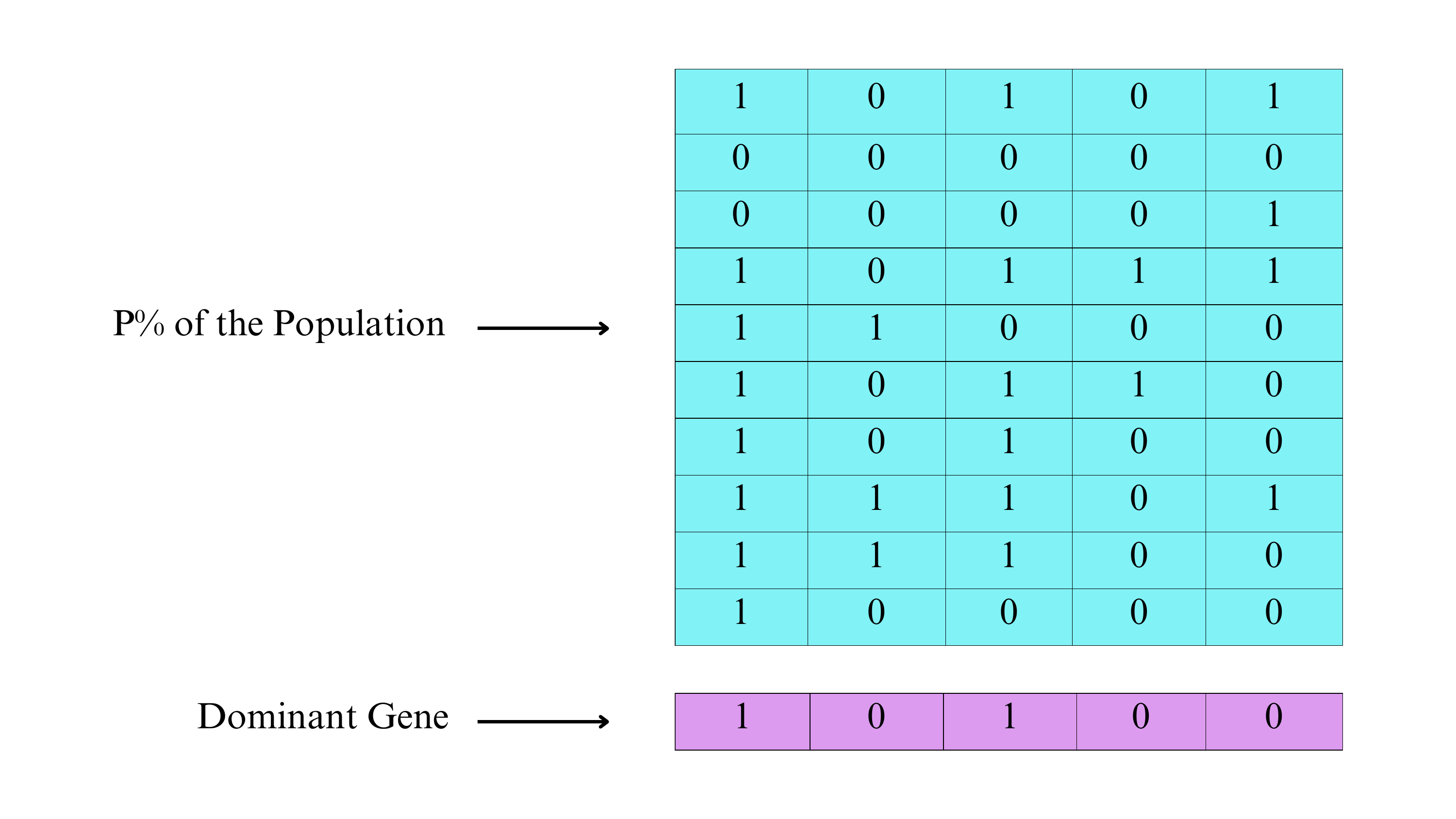}
\caption{Finding dominant genes from $p\%$ population.}
\end{figure}
\setlength{\textfloatsep}{0pt}

\underline{Senario 2.} 
\textbf{Directed Mutation:} The second scenario in GEA focuses on improving the effectiveness of the mutation operator, which is essential for preventing the algorithm from getting trapped in local optima. In traditional GAs, random mutation is often used, which can be a drawback. To address this limitation, the mutation operator in GEA is targeted and modified to enhance the efficiency of the genetic algorithm. In this scenario, specific methods are proposed to direct the mutation process, rather than relying on random selection. One such method involves the detection of desired genes, which enables the algorithm to focus on genes that are known to contribute to desired traits or outcomes. By targeting the mutation process and removing the randomness associated with traditional mutation, the performance of the genetic algorithm can be significantly improved. To provide an illustration of the directed mutation operator, Figure 3 presents an example that demonstrates how the process works. This targeted mutation approach allows the algorithm to prioritize specific genes and introduce beneficial changes in a controlled manner. By doing so, GEA can overcome the limitations of random mutation and enhance its ability to explore the solution space effectively.

\begin{equation}
    f(x)= \left\{ 
  \begin{array}{ c l }
    1 & \quad \textrm{if } M_{ij} \ desired \\
    0                 & \quad \textrm{otherwise}
  \end{array}
\right.
\end{equation}

\begin{itemize}
\item \textbf{Desired Gene:} The first step for applying this operator is to find the most repeated genes out of $p\%$ of the best chromosomes which are considered desired genes. The goal here is to consider fixed genes because these genes are considered the most informative elements for generating the elite part of the population. So, their existence in the solution will help the population to stay in the elite part, and by the slight change, they may move towards the global optimum in the near future. This step will generate a pattern matrix with $n*m$. where n is the number of populations in the elite part of the whole population (best $p\%$), and m is the number of genes inside a chromosome which is known as the number of the variables in the problem. This pattern matrix consists of binary elements, in which 1 represents the specific chromosome that is desired and should be fixed, and 0 represents uninformative genes for which mutation is allowed to be applied. Equation (3) represents how the pattern matrix will be generated. If the number of repetitions reaches a specific threshold then the gene will be desired. The threshold will be specified by the user from the beginning and the parameter can be optimized based on different problems and purposes.

\item \textbf{Mutation Targeting Desirable Genes:} After generating the pattern for $p\%$ chromosomes with the highest fitness values, a candidate by the roulette wheel will be selected, the mutation applies only on uninformative genes which represent zero in their corresponding pattern matrix. By applying targeting mutation we hope to search the solution space in the proper manner to find the global optimum. The engineered mutation will be repeated to the number of overall mutations only on uninformative genes to invest in the elite part of the population for faster convergence. 
\setlength{\textfloatsep}{0pt}
\end{itemize}

\underline{Senario 3.}
\textbf{Gene Injection:} The third scenario in GEA focuses on the importance of considering the entire population, including the individuals with the worst fitness values. While the first two scenarios primarily focus on the elite part of the population, it is essential to recognize that even the worst solutions can contribute to the overall improvement of the algorithm. In optimization algorithms, the worst individuals should not be overlooked, as they also possess the potential for beneficial changes. In this scenario, we aim to invest in the worst individuals and employ an engineering approach to enhance their performance. By making slight changes to the worst individuals, they have the opportunity to move towards the global optimum and eventually become part of the elite solutions in subsequent iterations of the algorithm.

To accomplish this, a patterns matrix is constructed for the elite part of the population. Then, individuals from the non-elite part of the population (representing $1-p\%$) are selected. Based on the pattern matrix, genes from the chromosome with the highest repetition are injected into the selected chromosome. This gene injection operator facilitates the transfer of beneficial genetic information from the dominant chromosome to other individuals in the population, enabling them to improve and contribute to the overall optimization process. Figure 4 provides an example that illustrates how this proposed gene injection operator works. It demonstrates the process of transferring genetic information to enhance the selected chromosome. The dominant chromosome, as coded in Algorithm 1, is necessary for the implementation of this gene injection operator.

By incorporating this third scenario into the GEA, we can harness the potential of even the worst solutions in driving the algorithm toward the global optimum. This approach allows for a more comprehensive exploration of the solution space and facilitates the improvement of the entire population over successive iterations.

\begin{figure}
\centering
\includegraphics[width=1\columnwidth]{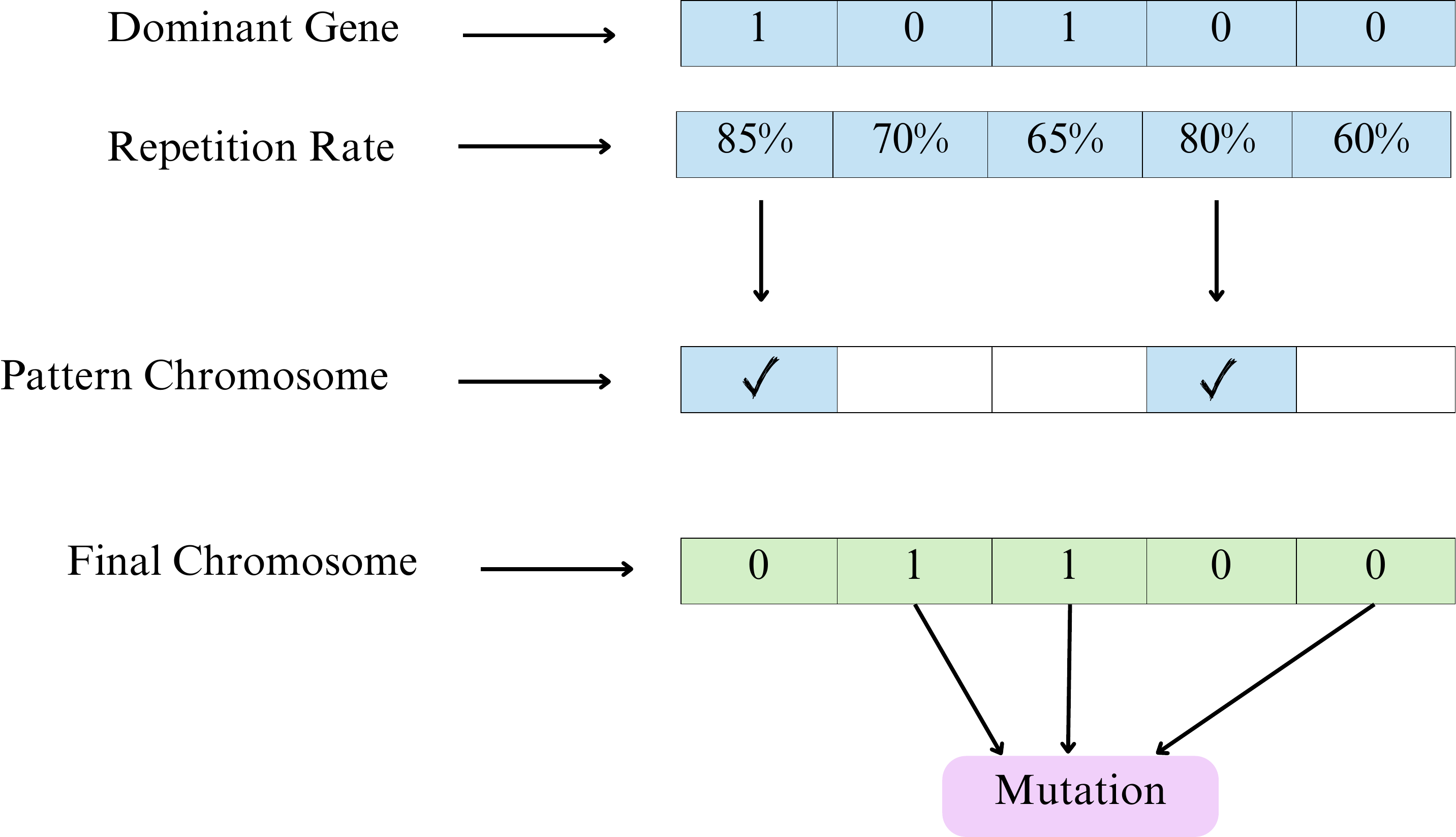}
\caption{Directed mutation by fixing informative genes.}
\end{figure}
\setlength{\textfloatsep}{0pt}

\begin{algorithm}[H]
\caption{Dominant Chromosome}
\textbf{Data: }{Pop, Prob. Info.} \\
\textbf{Result: }{DominantGene, Mask, MaskInverted} \\ \\
 \While{i less than chromosome length}{
 \While{j less than No. of Pop.}{
    $Genes\gets [Genes, Pop_j(i)]$
 }
 \While{there is element in Genes}{
    $temp\gets sum(Genes == Genes(1))$ \\
 \eIf{size DominantGene == 0}{
    $DominantGene\gets Genes(1)$ \\
    $DominantGeneCounter\gets temp$
 }{
 \eIf{temp > DominantGeneCounter}{
    $DominantGene\gets Genes(1)$ \\
    $DominantGeneCounter\gets temp$
 }{$DominantGene\gets [DominantGene, Genes(1)]$}
 }
 }
 }
$Mask\gets zeros(1, size(chromosome))$ \\
\While{i less than chromosome length}{
\If{(DominantGeneCount > threshold) and (threshold not 0)}{
    $Mask(i)\gets 1$
}
}
    $MaskInverted\gets not Mask$
\end{algorithm}

\setlength{\textfloatsep}{0pt}
\begin{figure}
\centering
\includegraphics[width=1\columnwidth]{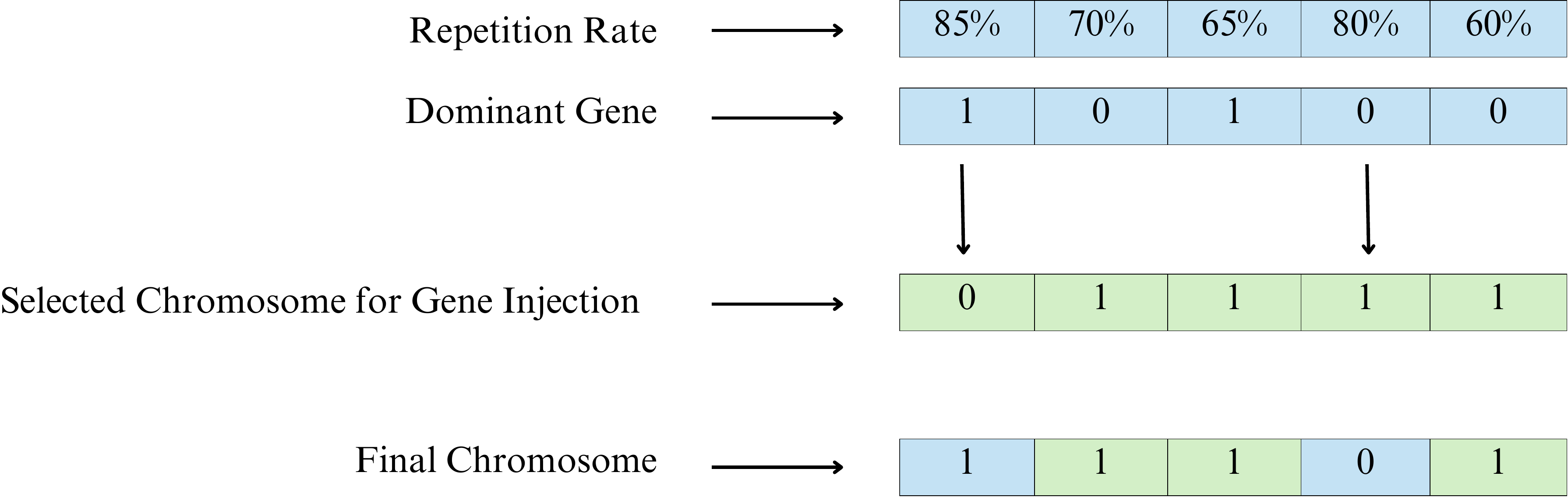}
\caption{Injecting informative genes to wort individuals of the population.}
\end{figure}

\section{Experimental Results}

Here, we present a comprehensive evaluation of the GEA and demonstrate its efficacy in solving combinatorial optimization problems, particularly in the context of a standard vehicle routing optimization problem. This problem involves determining optimal routes for a fleet of vehicles to visit a set of demand points while minimizing transportation costs. To evaluate the performance of GEA, we compare it not only against the traditional GA but also against three variations of GEA, namely GEA1, GEA2, and GEA3, each utilizing a specific scenario as explained earlier. In GEA, the main loop randomly selects one of these scenarios at each iteration.

For our evaluation, we select six well-established instances from the literature, as referenced by ~\cite{ref18} and ~\cite{ref9}, to benchmark the algorithms. To ensure consistency, we set the maximum number of iterations to 1000 and the population size to 100 for all algorithms. The crossover and mutation percentages are uniformly set to 0.8 and 0.1, respectively, across all algorithms. Moreover, in the case of GEA, the percentages of scenarios considered are 0.5, 0.5, and 0.2 for the first, second, and third scenarios, respectively.

\begin{table}
\centering
\captionof{table}{Report of the algorithms results based on criteria of the Best=B, Worst=W, Mean=M, and Standard deviation=Std. (The best values in each criterion and test instance are highlighted in bold.)} \label{tab:title}

\begin{tabular}{|l|l|l|l|l|l|l|l|} 
\hline
\multicolumn{2}{|l|}{Test instance}                                                                & F1                & F2                & F3                & F4                & F5                & F6                 \\ 
\hline
\multicolumn{2}{|l|}{\begin{tabular}[c]{@{}l@{}}Demand points x\\Number of vehicle\end{tabular}}   & 8x3               & 10x3              & 14x4              & 20x4              & 25x5              & 30x5               \\ 
\hline
\multirow{4}{*}{GA}     & B                                                                        & \textbf{257.3492} & \textbf{268.1687} & \textbf{301.6661} & 317.6503          & 326.5457          & 308.8542           \\ 
\cline{2-2}
                        & W                                                                        & 291.6624          & 269.0742          & 316.0882          & 351.2298          & 363.0131          & 343.9097           \\ 
\cline{2-2}
                        & M                                                                        & 260.7805          & 268.7120           & 305.8615          & 333.5178          & 338.1742          & 321.1532           \\ 
\cline{2-2}
                        & Std                                                                      & 10.8507          & 0.4675          & 5.4002          & 12.3733          & 11.3976          & 11.5798           \\ 
\hline
\multirow{4}{*}{GEA\_1} & B                                                                        & \textbf{257.3492} & \textbf{268.1687} & \textbf{301.6661} & 319.3303          & 319.5602          & 307.1991           \\ 
\cline{2-2}
                        & W                                                                        & \textbf{257.3492} & 269.0742          & 318.8057          & 342.2278          & 359.0854          & 370.7047           \\ 
\cline{2-2}
                        & M                                                                        & \textbf{257.3492} & 268.2593          & 304.7409          & 324.0378          & 330.8722          & 328.0479           \\ 
\cline{2-2}
                        & Std                                                                      & \textbf{5.99E-14} & 0.2863          & 5.4787          & 6.8149          & 10.8851          & 21.2861           \\ 
\hline
\multirow{4}{*}{GEA\_2} & B                                                                        & \textbf{257.3492} & \textbf{268.1687} & \textbf{301.6661} & \textbf{317.1235} & 321.5556          & \textbf{302.5377}  \\ 
\cline{2-2}
                        & W                                                                        & \textbf{257.3492} & 269.0742          & 306.3834          & 353.7992          & 359.0854          & \textbf{322.7266}  \\ 
\cline{2-2}
                        & M                                                                        & \textbf{257.3492} & 268.3498          & 302.8296          & 327.4803          & 333.1713          & \textbf{311.4745}  \\ 
\cline{2-2}
                        & Std                                                                      & \textbf{5.99E-14} & 0.3817          & 1.6555          & 12.1645          & 13.4915          & \textbf{7.3910}  \\ 
\hline
\multirow{4}{*}{GEA\_3} & B                                                                        & \textbf{257.3492} & \textbf{268.1687} & \textbf{301.6661} & 317.6503          & 319.0169          & 308.8834           \\ 
\cline{2-2}
                        & W                                                                        & \textbf{257.3492} & 269.0742          & 306.3834          & \textbf{331.8416} & \textbf{331.3571} & 346.2497           \\ 
\cline{2-2}
                        & M                                                                        & \textbf{257.3492} & 268.4404          & 302.3684          & 323.3476          & 326.245           & 323.5826           \\ 
\cline{2-2}
                        & Std                                                                      & \textbf{5.99E-14} & 0.4373          & 1.58597          & 5.45505          & \textbf{4.1059} & 13.7657           \\ 
\hline
\multirow{4}{*}{GEA}    & B                                                                        & \textbf{257.3492} & \textbf{268.1687} & \textbf{301.6661} & 317.6503          & \textbf{317.7347} & 304.4598           \\ 
\cline{2-2}
                        & W                                                                        & \textbf{257.3492} & \textbf{268.1687} & \textbf{301.6661} & \textbf{331.8416} & 331.4877          & 343.6004           \\ 
\cline{2-2}
                        & M                                                                        & \textbf{257.3492} & \textbf{268.1687} & \textbf{301.6661} & \textbf{321.4611} & \textbf{323.1822} & 313.4242           \\ 
\cline{2-2}
                        & Std                                                                      & \textbf{5.99E-14} & \textbf{5.99E-14} & \textbf{~ ~ ~0}   & \textbf{4.7726} & 6.0097          & 11.8294           \\
\hline
\end{tabular}
\end{table}

\begin{figure}
     \centering
     \begin{subfigure}[b]{0.48\textwidth}
         \centering
         \includegraphics[width=\textwidth]{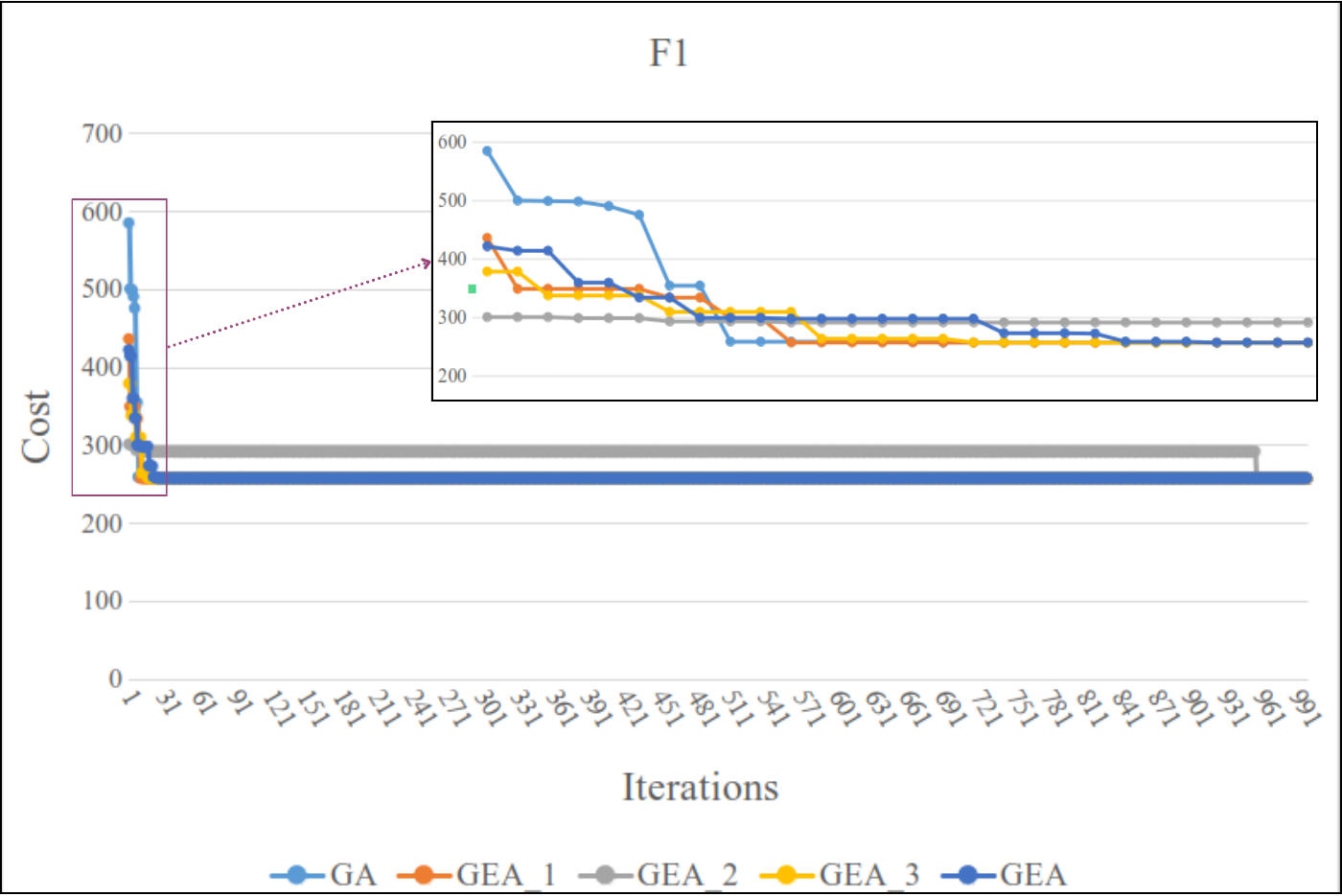}
     \end{subfigure}
     \hfill
     \begin{subfigure}[b]{0.48\textwidth}
         \centering
         \includegraphics[width=\textwidth]{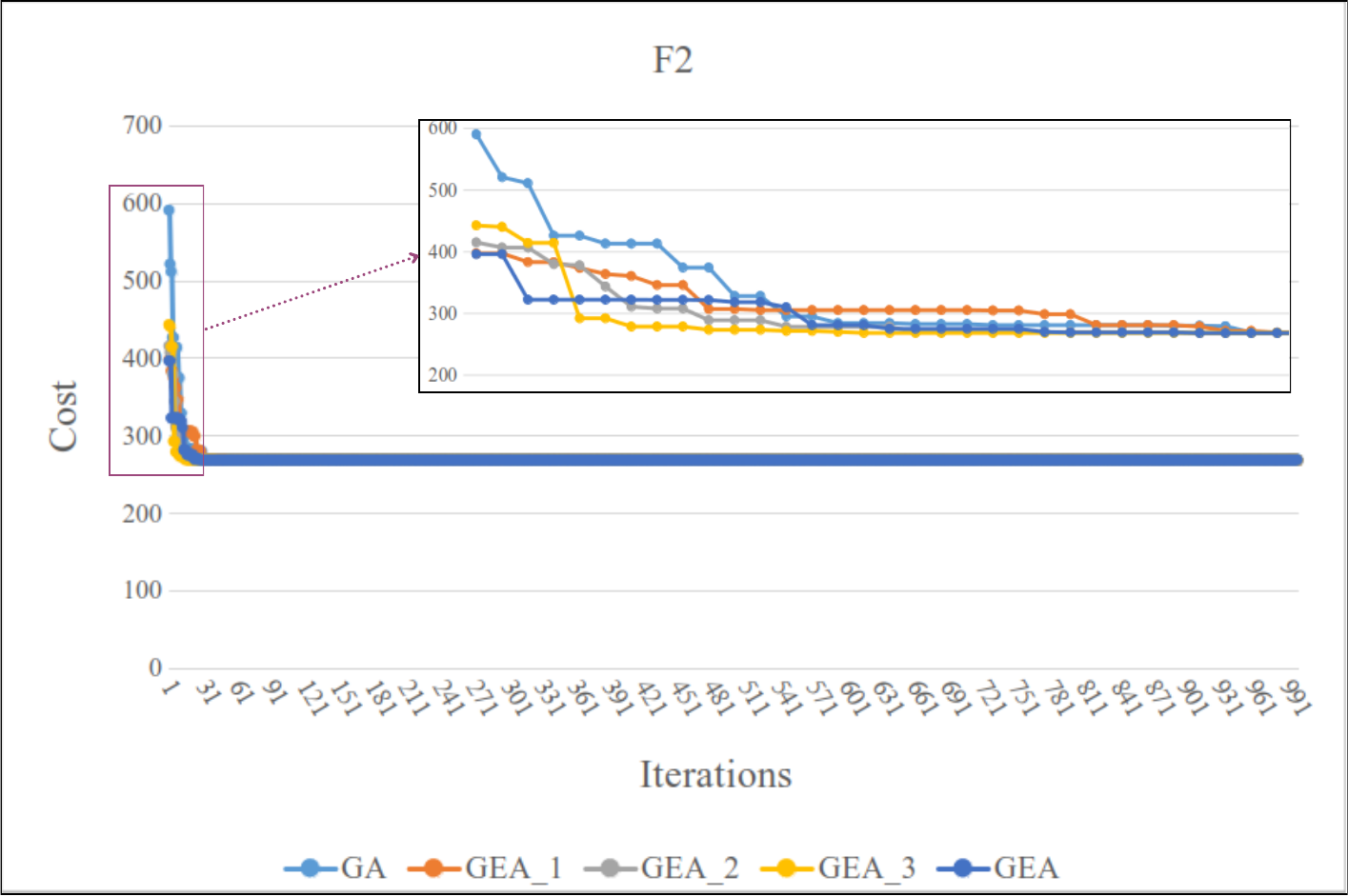}
     \end{subfigure}
     \hfill

     \centering
     \begin{subfigure}[b]{0.48\textwidth}
         \centering
         \includegraphics[width=\textwidth]{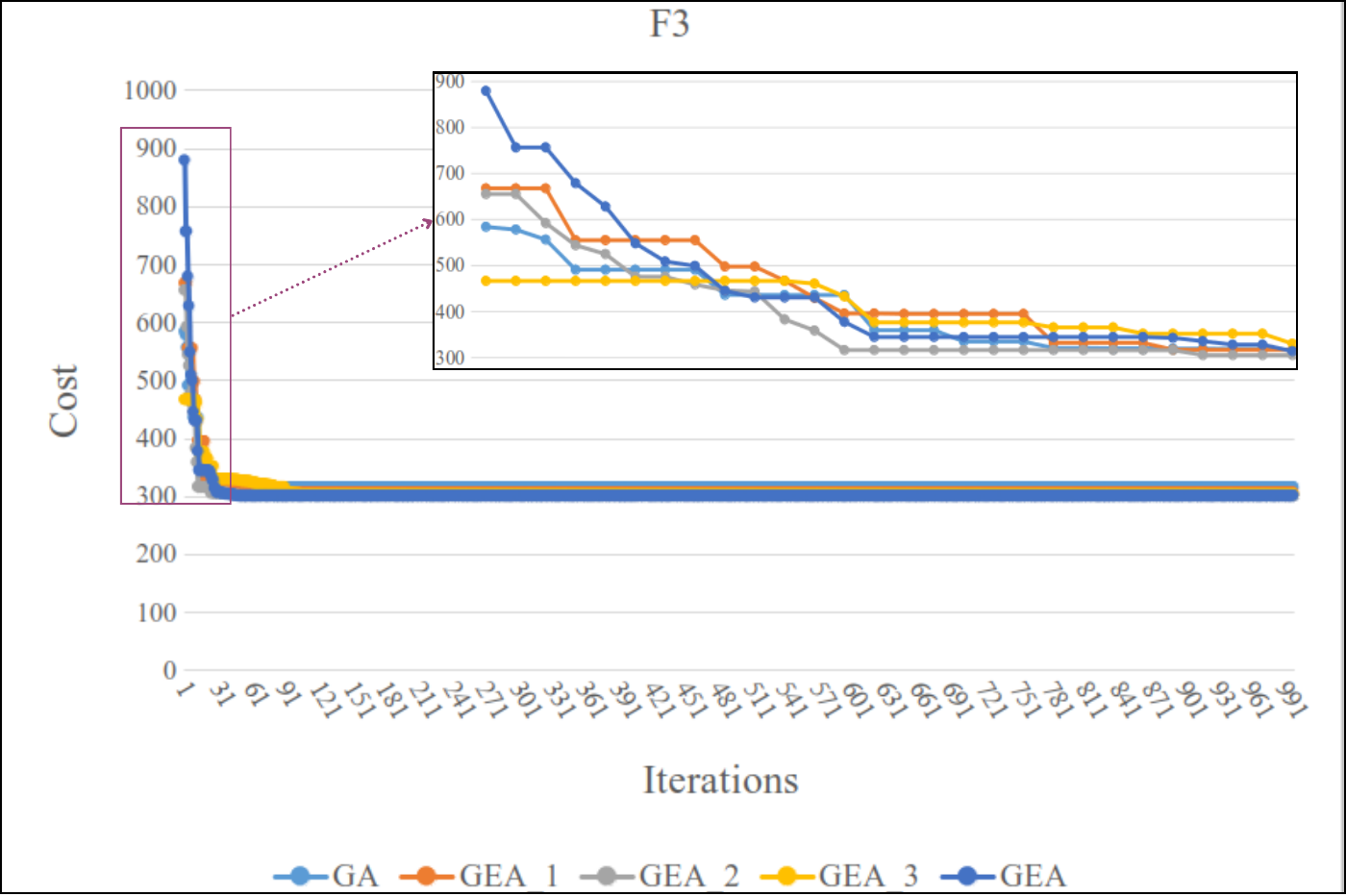}
     \end{subfigure}
     \hfill
     \begin{subfigure}[b]{0.48\textwidth}
         \centering
         \includegraphics[width=\textwidth]{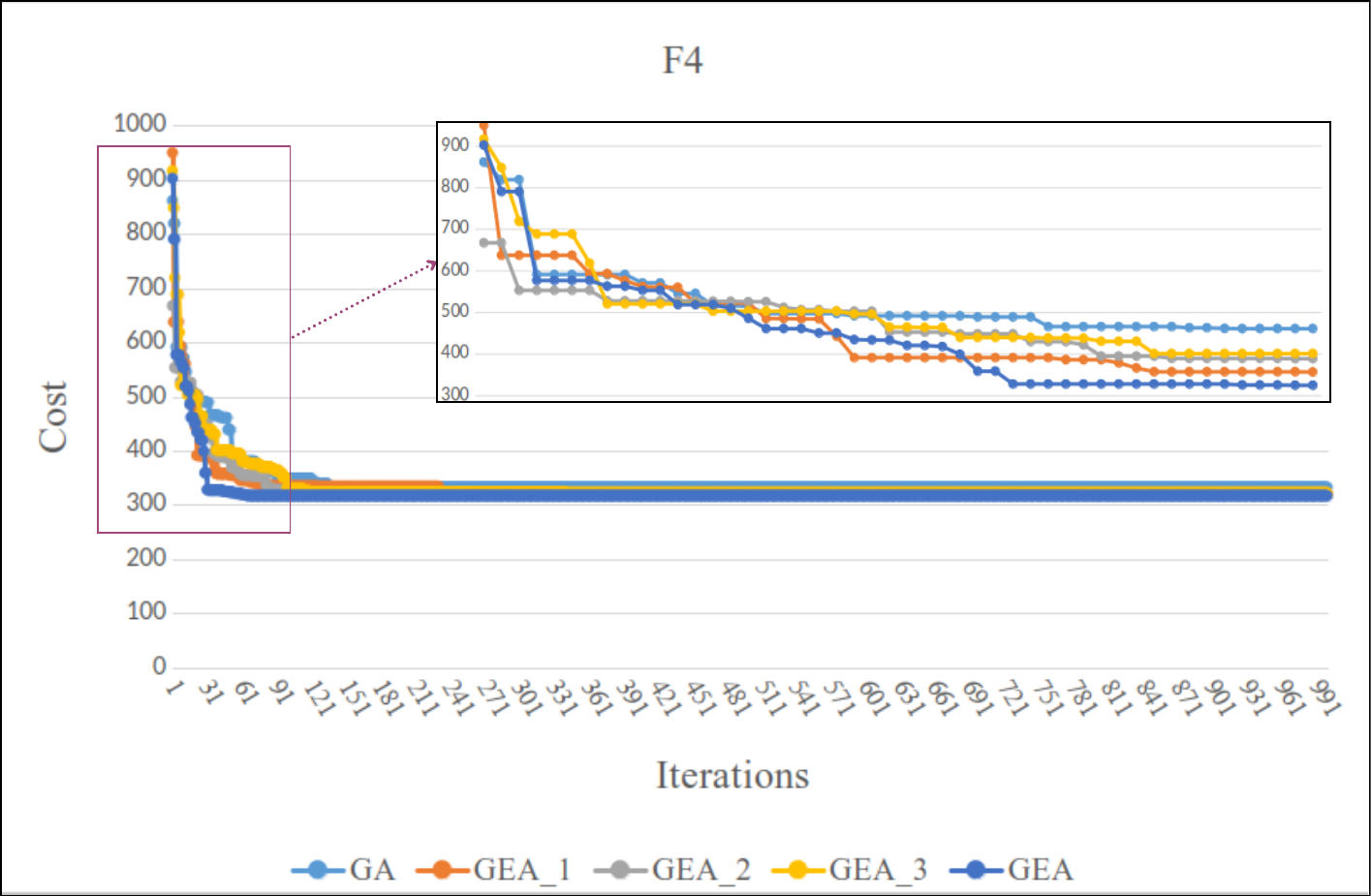}
     \end{subfigure}
     \hfill

     \centering
     \begin{subfigure}[b]{0.48\textwidth}
         \centering
         \includegraphics[width=\textwidth]{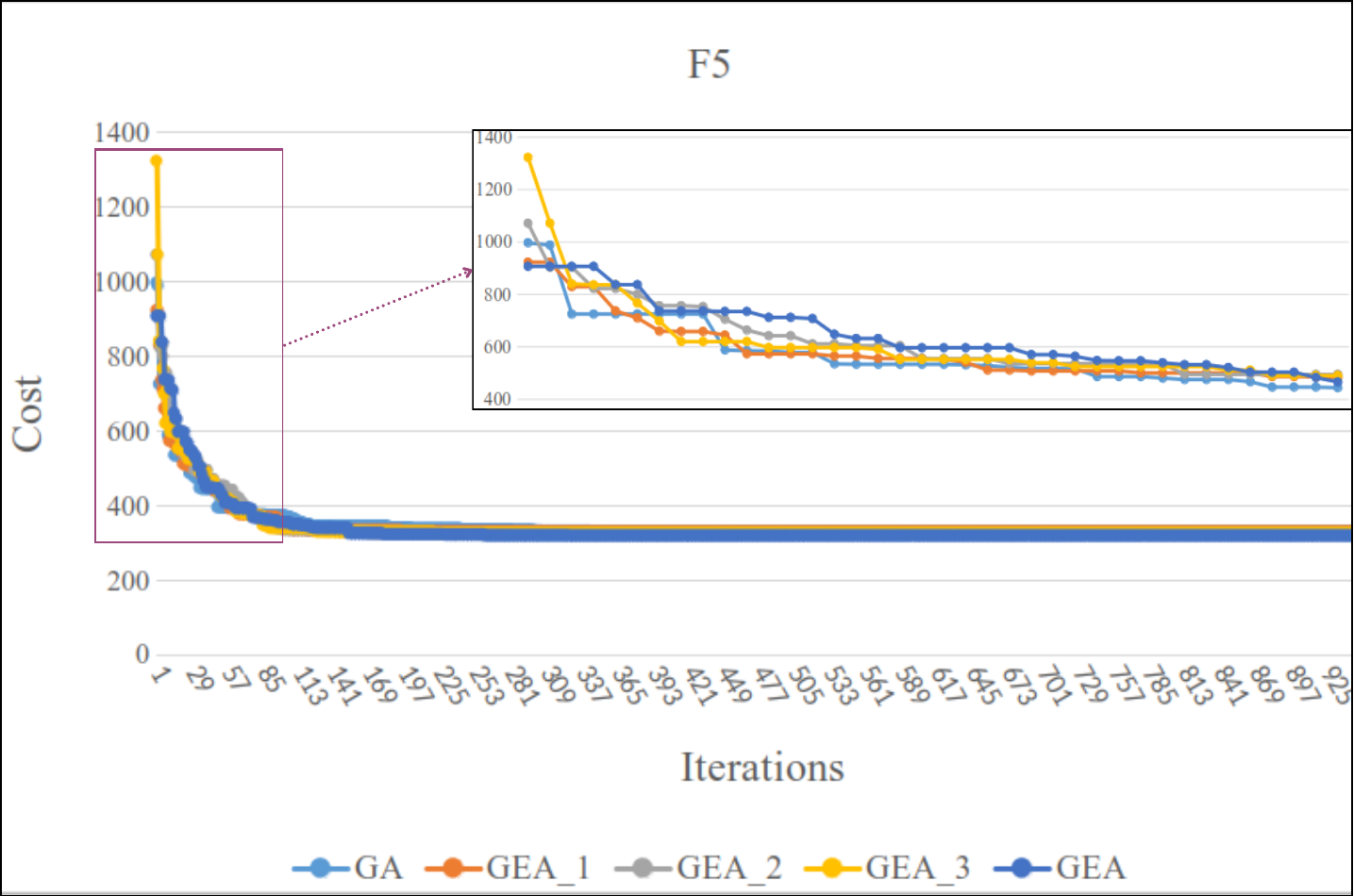}
     \end{subfigure}
     \hfill
     \begin{subfigure}[b]{0.48\textwidth}
         \centering
         \includegraphics[width=\textwidth]{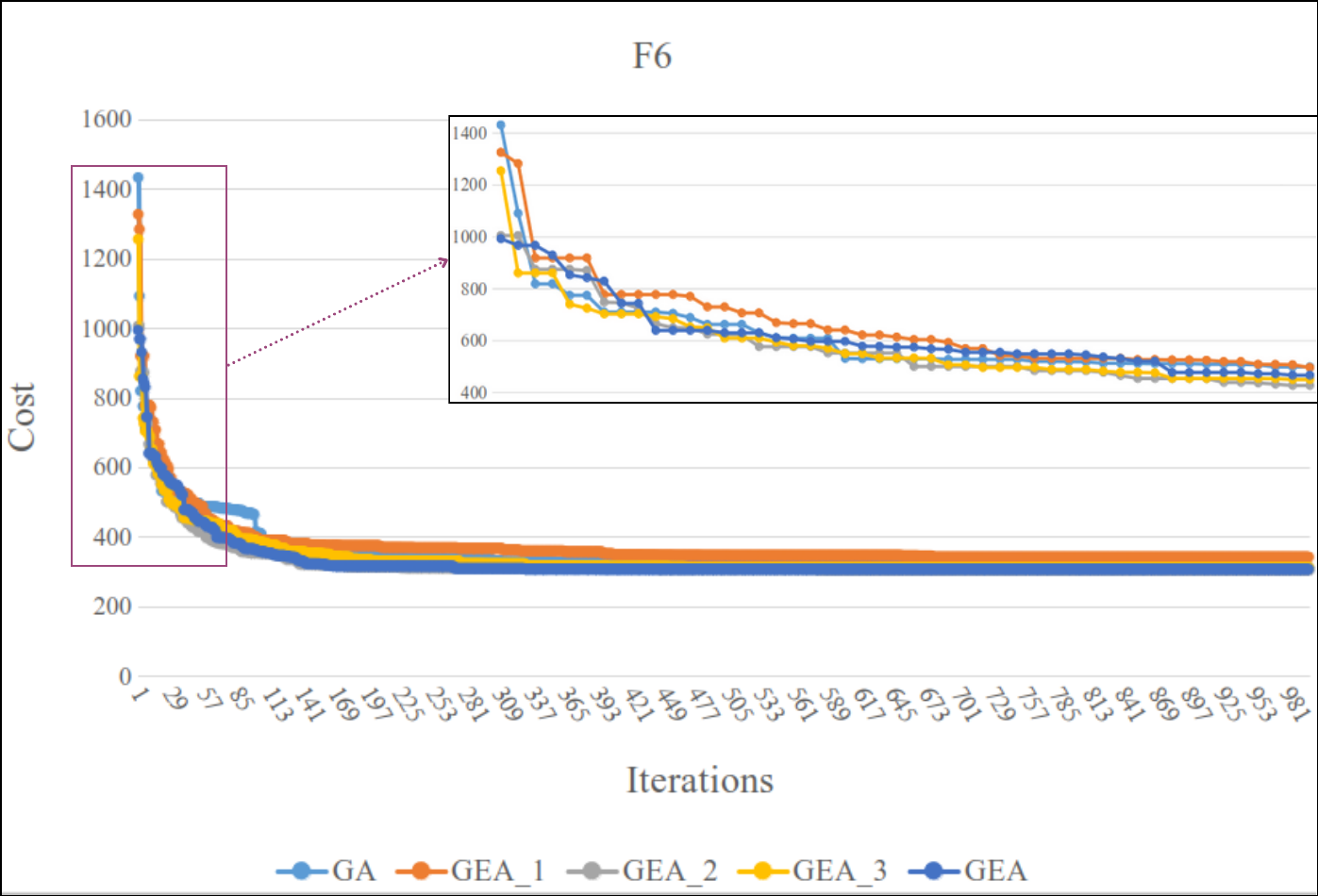}
     \end{subfigure}
     \hfill
\caption{Convergence rate of the metaheuristic algorithms in all the benchmarked instances.}
\end{figure}

\begin{figure}
\centering
\includegraphics[width=0.6\columnwidth]{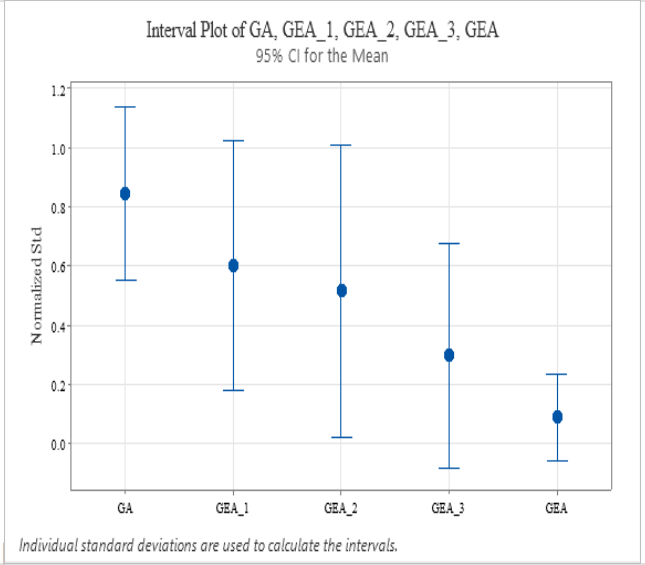}
\caption{Interval plot based on $95\%$ confidence level for analyzing the robustness of the metaheuristic algorithms}
\end{figure}

To gauge the performance of the algorithms, we conduct 10 independent runs of each algorithm on every test instance. Subsequently, we report the best, worst, average, and standard deviation of the solutions obtained by each algorithm in Table 1. These results enable us to analyze the robustness of the applied metaheuristic algorithms. Furthermore, Figure 5 illustrates the convergence rate of each algorithm towards its best performance, providing a visual representation of their effectiveness. To complement our analysis, we perform statistical analyses using a 0.95 confidence level, employing normalized standard deviations across all algorithms. The interval plot in Figure 6 showcases the results of these statistical analyses, shedding light on the comparative performance and reliability of the algorithms.

Based on the results presented in Table 1, our findings indicate that GEA, when utilizing all scenarios, outperforms the other algorithms. In most instances, this algorithm consistently discovers near-optimal solutions superior to those obtained by GA and the other GEA variants. Among the GEA variations, GEA2 stands out as the most successful, confirming the strength of the second scenario in exploring better near-optimal solutions. Figure 5 demonstrates that all algorithms exhibit an acceptable convergence rate across the test instances, with similar solution quality. However, the statistical analyses from Figure 6 conclusively support the highest accuracy of our GEA compared to the other algorithms.

In conclusion, our comprehensive evaluation showcases the effectiveness of the GEA in solving combinatorial optimization problems, specifically the vehicle routing optimization problem. The results presented in Table 1 and Figures 5 and 6 highlight the superior performance and accuracy of GEA, particularly when incorporating all scenarios. These findings provide compelling evidence for the potential of GEA as a robust and reliable metaheuristic algorithm for addressing optimization challenges.

\section{Conclusion and Future Work}

This study presented a comprehensive evaluation of the GEA and its effectiveness in solving combinatorial optimization problems, focusing on the vehicle routing optimization problem. The results obtained through benchmarking against the traditional GA and different variations of GEA demonstrated the superiority of GEA, particularly when incorporating all scenarios. GEA consistently outperformed other algorithms, yielding better near-optimal solutions in most instances.

The findings of this study contribute to the growing body of research on metaheuristic algorithms and their applications in optimization problems. The success of GEA in addressing the vehicle routing optimization problem showcases its potential in real-world scenarios where efficient transportation routing is critical. These results provide valuable insights into the efficacy of genetic engineering techniques based on our three scenarios for the development of GEA in solving combinatorial optimization problems and highlight the importance of considering different scenarios in the algorithm design.

Moving forward, several areas for future research can be identified. Firstly, further investigation can be conducted to explore the impact of different parameter settings on the performance of GEA and its variations. Fine-tuning the algorithm's parameters may enhance its effectiveness and lead to even better solutions. Extending the evaluation to other optimization problems and comparing GEA against state-of-the-art algorithms would provide a broader perspective on its performance and competitiveness. Moreover, integrating GEA with other optimization techniques or hybridizing it with machine learning approaches could further enhance its capabilities. Combining the strengths of genetic engineering algorithms with other intelligent algorithms may lead to novel hybrid approaches with improved optimization performance. Lastly, conducting experiments on larger problem instances and assessing the scalability and efficiency of GEA would be beneficial ~\cite{ref40}. Scaling up the problem size would provide insights into the algorithm's performance when dealing with more complex and larger-scale optimization challenges ~\cite{ref41}.

\subsubsection{Acknowledgements} This paper is an output of a research project implemented as part of the Basic Research Program at the National Research University Higher School of Economics (HSE University). We gratefully acknowledge the HPC facilities at HSE University~\cite{ref27} for providing computational resources. In addition, we would like to acknowledge that the publications from Prof. Mostafa Hajiaghaei-Keshteli inspire us to develop a new metaheuristic algorithm.

\end{document}